\pdfoutput=1

\documentclass[11pt]{article}

\usepackage[final]{acl}

\usepackage{times}
\usepackage{latexsym}

\usepackage[T1]{fontenc}

\usepackage[utf8]{inputenc}

\usepackage{microtype}

\usepackage{inconsolata}

\usepackage{graphicx}
\usepackage{times}
\usepackage{xspace}
\usepackage{latexsym}
\usepackage{booktabs}
\usepackage{comment}
\usepackage{tcolorbox}
\usepackage{enumitem}

\usepackage{tikz}

\newcommand*{\warri}{\textsc{Warri}\xspace}
\newcommand*{\mafand}{\textsc{Mafand}\xspace}
\newcommand*{\ud}{\textsc{UD}\xspace}
\newcommand*{\jw}{\textsc{JW300}\xspace}
\newcommand*{\bible}{\textsc{Bible}\xspace}
\newcommand*{\bbc}{\textsc{BBC}\xspace}
\newcommand*{\wiki}{\textsc{Wiki}\xspace}
\newcommand*{\gpt}{\textsc{GPT-4o}\xspace}
\newcommand*{\llama}{\textsc{LLaMa 3.1 8B}\xspace}
\newcommand*{\llamaB}{\textsc{LLaMa 3.1 70B}\xspace}
\newcommand*{\yoruba}{Yor\`ub\'a\xspace}

\newcommand\ChapterPrecis[2]{%
\begin{tikzpicture}[remember picture,overlay]
\node[anchor=north, draw=black, fill=yellow!20, inner sep=5pt, rounded corners, align=left, yshift=-#1] at (current page.north) 
{\parbox[t][1.2cm][l]{16cm}{\small #2}};
\end{tikzpicture}%
}

\title{Does Generative AI speak Nigerian-Pidgin?: \\
Issues about Representativeness and Bias for Multilingualism in LLMs}

\author{First Author \\
  Affiliation / Address line 1 \\
  Affiliation / Address line 2 \\
  Affiliation / Address line 3 \\
  \texttt{email@domain} \\\And
  Second Author \\
  Affiliation / Address line 1 \\
  Affiliation / Address line 2 \\
  Affiliation / Address line 3 \\
  \texttt{email@domain} \\}

\author{David Ifeoluwa Adelani$^*$ \\ Mila - Quebec AI Institute \\ McGill University, Canada CIFAR AI Chair\\ \texttt{david.adelani@mcgill.ca} \\ \And
A. Seza Doğruöz\thanks{Equal contribution} \\ LT3, IDLab, Universiteit Gent \\\texttt{as.dogruoz@ugent.be} 
\\ \AND
Iyanuoluwa Shode\\  Bloomberg \\ \texttt{ishode@bloomberg.net} \\ \And
        Anuoluwapo Aremu \\ 
        University of Trento \ \\
Lelapa AI \\  }

\begin{document}

\maketitle

  \ChapterPrecis{1.1cm}{If you cite this paper, \textbf{please use the official ACL Anthology reference:} \\
  David I. Adelani, A. Seza Doğruöz, Iyanuoluwa Shode, and Anuoluwapo Aremu. 2025. \href{https://aclanthology.org/2025.findings-naacl.85/}{Does Generative AI speak Nigerian-Pidgin?: Issues about Representativeness and Bias for Multilingualism in LLMs}. In Findings of the Association for Computational Linguistics: NAACL 2025, pages 1571–1583, Albuquerque, New Mexico. Association for Computational Linguistics. \color{blue}{\url{}}
}

\begin{abstract}

Nigeria is a multilingual country with 500+ languages. Naija is a Nigerian Pidgin spoken by approximately 120M speakers and it is a mixed language (e.g., English, Portuguese, Yoruba, Hausa and Igbo). Although it has mainly been a spoken language until recently, there are some online platforms (e.g., Wikipedia), publishing in written Naija as well. West African Pidgin English (WAPE) is also spoken in Nigeria and it is used by BBC to broadcast news on the internet to a wider audience not only in Nigeria but also in other West African countries (e.g., Cameroon and Ghana). Through statistical analyses and Machine Translation experiments, our paper shows  that these two pidgin varieties do not represent each other (i.e., there are linguistic differences in word order and vocabulary) and Generative AI operates only based on WAPE. In other words, Naija is underrepresented in Generative AI, and it is hard to teach LLMs with few examples. In addition to the statistical analyses, we also provide historical information on both pidgins as well as insights from the interviews conducted with volunteer Wikipedia contributors in Naija.  

\end{abstract}

\section{Introduction}
Between 16th-19th centuries, there were contacts between Europeans and non-Europeans outside Europe. In West Africa, contacts between English and West African languages led to simplified and mixed languages combining linguistic features from several languages. These new forms of languages were lingua francas (i.e., common or bridge languages) that served for a mutual understanding between speakers of different languages for various purposes (e.g., trade, plantation agriculture, mining)~\citep{mufwene2024pidgin}. The terms "pidgin" and "creole" are used to refer to these languages. Although there is a lack of agreement about the precise definitions and coverage of these terms, pidgin roughly refers to the "speech-forms which do not have native speakers, and are therefore primarily used as a means of communication among people who do not share a common language" \cite{muysken1995study}. Creoles, on the other hand, are assumed to be extended pidgins which are more established and have native speakers especially in urban environments \cite{muysken1995study}.

Nigeria is a multilingual country in West Africa hosting over 500 different languages spoken by approximately 220 million people across 371 ethnic tribes \cite{eberhard2019ethnologue}. It is the sixth most populous country in the world and Africa’s most populous country. 
English is the official language and acquired mostly through formal education in Nigeria \cite{agbo2020relationship}. The three major tribes in Nigeria with their respective languages include Hausa (spoken by 63M speakers), Igbo (27M speakers), and \yoruba (42M speakers). Nigerian Pidgin (Naija) is a mix of English with local languages (e.g., Portuguese, \yoruba, Igbo, Hausa) \cite{balogun2013defense}, \cite{oyebola2023attitudes}. Naija is widely spoken (approx. 120M speakers) as a first and second language \cite{Adelani_2022} around the Southern part of Nigeria (e.g., Lagos and  Niger-Delta) with origins going back to the English-Creole Atlantic Krio language family. It is also adopted as the unifying and unofficial language for communication across ethnically diverse groups. According to some researchers (e.g., \newcite{muysken1995study}), Naija has evolved into a creole over time and has now native speakers as well. However, it is still  referred to as a pidgin among the locals.

Although Naija may have words that sound similar to English, their meanings may vary and there is no standardized orthography  for Naija ~\citep{marchal-etal-2021-semi,akande_book,lin2024modeling}. 






Until recently,\textbf{ West African Pidgin English (WAPE)} has been mainly a spoken language with many local varieties (e.g., Nigerian Pidgin, Ghanaian Pidgin English, Cameroonian Pidgin English). Despite the large number of speakers across West Africa countries, WAPE remained as a spoken language until 2017 when the British Broadcasting Company (BBC) launched a news website (West African Pidgin English). It aims to target the diversity of WAPE speakers across different countries.~\footnote{\url{https://www.bbc.com/news/world-africa-40975399}} Since 2022, Naija is also accepted as one of the languages on Wikipedia~\footnote{\url{https://meta.wikimedia.org/wiki/Requests_for_new_languages/Wikipedia_Nigerian_Pidgin}}. Although they are mutually understandable, there are linguistic and social differences between the two written varieties. For example, (WAPE on BBC website) resembles English in terms of word order and vocabulary (see example (1)) with a simplified grammar, lacking auxiliary "were"). However, Naija on Wikipedia has a different word order and vocabulary choice (e.g., ''moto'' instead of a "car" and ''wund" instead of ''injured" or ''wounded"). 

\begin{figure}[t]
    \centering
    \includegraphics[width=1.02\linewidth]{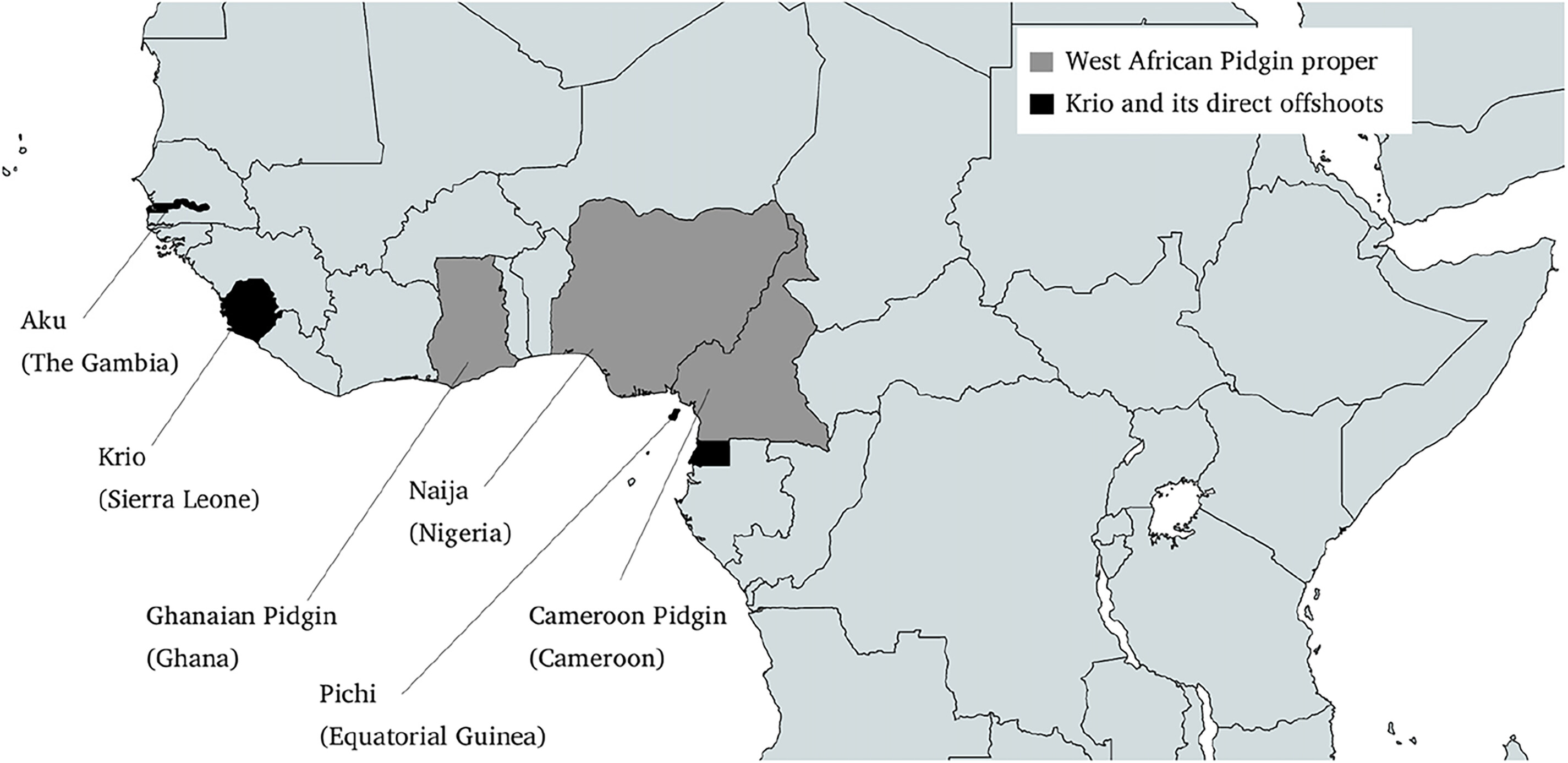}
    \caption{\textbf{WAPE Locations}: West African countries whose official language is English (The Gambia, Sierra Leone, Liberia, Ghana and Nigeria), Cameroon (North-Western and South-Western Anglophone region), and Krios immigrants to Equatorial Guinea. Map is obtained from \citet{kofi_wape}. }
    \label{fig:transfer_en_pcm}
\end{figure}

\textbf{Example (1)}
\begin{enumerate}
    \item[] \textbf{WAPE} Two pesin in di car dey injured.
    \item[] \textbf{Naija} Na wund di two pesin get for di moto.
    \item[] \textbf{English Translation:} Two persons in the car were injured. 
\end{enumerate}

From the sociolinguistic perspective, WAPE is favored mostly by educated Nigerians (who also speak, read and write in English) whereas Naija is used mostly in everyday life and it is more accessible for a larger audience ~\citep{Akande2012UseAA}. In terms of the resources, there is much more data on the Internet about WAPE than Naija (160K sentences vs. 25K sentences)~\citep{ogueji-etal-2021-small}.


There is a growing need for more research for multilingual and low resource languages \cite{dogruoz-sitaram-2022-language, dogruoz-etal-2021-survey} in Generative AI systems. This need is even enhanced for pidgin and/or creole languages due to their high numbers of speakers but lack of data \cite{lent-etal-2022-creole, Lent2023CreoleValMM}. However, there are also unresolved issues about to what extent the available data on the Internet represent the language spoken in real life \cite{dogruoz-etal-2023-representativeness}. 
It is crucially important that the different genres of the same pidgin and creole languages are also represented in these systems to be inclusive and accessible for all speakers/users with diverse backgrounds. 

In our paper, we address these issues for WAPE and Naija in the Nigerian context with the following contributions. We introduce \warri~\footnote{Warri is the name of a city in Delta State in the South-South region of Nigeria where Naija is widely spoken} as a new MT evaluation data set including written WAPE on the BBC website and written Naija on Wikipedia. We also had interviews with the Naija Wikipedia contributors to understand their motives and regulations about written Naija. 
Our paper is the first to systematically analyze the similarities and differences between WAPE and Naija in Nigeria. Through a Machine Translation (MT) experiment, we find that Generative AI models (e.g., \gpt and \llama~\citep{Touvron2023Llama2O}) are biased towards WAPE  and they do not include Naija despite large numbers of speakers. Further analysis shows that LLMs are hard to teach with few examples (e.g., 5-shots) to generate text in Naija. For reproducibility purposes, we release the Warri dataset and our evaluation code on GitHub under CC-BY-4.0 license.\footnote{\url{https://github.com/McGill-NLP/Naija-representation-in-LLMs}}

\section{Related Research}
Available research on representativeness originates from corpus linguistics where it is important to include samples from different textual sources to have a balanced and representative (smaller) corpus reflecting the variation in the (larger) corpora \cite{Biber1993RepresentativenessIC}. 
Similarly, \newcite{Crowdy1993SpokenCD} states the significance of representative sampling corpora to minimize bias and maximize the credibility and consistency of the linguistic analyses. Therefore, representative sampling encompasses a broad spectrum of language usage across various contexts, genres, and demographic factors. 

Generative AI systems depend on the availability of large data sets on the Internet. However, this assumption does not consider the representativeness of the variation in the available data sets which is especially difficult to obtain for multilingual and low resource languages \cite{dogruoz-etal-2023-representativeness}. While developing language technologies for multilingual and low resource languages, it is crucially important to be aware of the linguistic variation (e.g., WAPE and Naija) within these languages and aim for representing the variation in a balanced way to prevent potential bias. 

To investigate the potential bias in the Nigerian context, the first step is to establish to what extent WAPE and Naija are similar or different from each other linguistically.

\section{\warri MT benchmark dataset 
for WAPE and Naija}
\begin{table*}[t]
 \begin{center}
 \footnotesize
  \begin{tabular}{l|llrrrr}
    \toprule
    \textbf{Dataset} & \textbf{Creole} & \textbf{Domain} & \textbf{Average length (Pidgin)} & \textbf{TRAIN} & \textbf{DEV} & \textbf{TEST}\\
    \midrule
Bible & PCM & religious & 25.2 & 31,051 & 1,500 & 1,500 \\
JW300 & PCM & religious & 17.2 & 23,322 & 1,500 & 1,500 \\
UD & PCM & spoken & 10.6 & 6241 & 1,500 & 1,500 \\
\mafand & PCM & news & 25.0 & 4,790 & 1,484 & 1, 564 \\
\midrule
\warri (single-way) & WAPE & news (BBC) & 20.8 & 5& - & 500 \\
\warri (multi-way) & WAPE \& PCM & Wiki & 21.3 & 5& - & 500  \\

    \bottomrule
  \end{tabular}
  \vspace{-2mm}
  \caption{\textbf{\warri and other datasets:} \warri is only used for evaluation in zero or few-shot (e.g. 5) setting. \warri (multi-way) have the same sentences in both WAPE and Naija (PCM is the ISO 639-3 code) pidgins unlike \warri (single-way). We label each dataset based on the specified pidgin assigned by the creators of the dataset.} 
  \label{tab:dataset}
  \end{center}
  \vspace{-3mm}
\end{table*}
To establish the \warri data set, we used a portion of WAPE BBC news data, previously used in MasakhaNER dataset~\citep{adelani-etal-2021-masakhaner}. It is a Named Entity Recognition (NER) dataset with available untokenized texts. We downloaded the Naija Wikipedia data from the Hugging Face.\footnote{We make use of the 20231101 version, \url{https://huggingface.co/datasets/wikimedia/wikipedia}} After the data collection, we created a parallel data set in English by recruiting two bilingual speakers.  They translated about $505$ sentences from WAPE BBC data and Naija Wikipedia data into English. In this way, we maintained  high-quality datasets by preventing  the translators from mixing the features of the two pidgins into one. However, this also introduced a new obstacle (i.e., comparison of two test sets from slightly different domains (news vs. Wikipedia)).

To handle the domain related obstacle, we created a \textbf{multi-way parallel dataset} for Wikipedia domain. First, we asked a bilingual speaker to translate the Naija Wikipedia sentences into English. Then, we asked a professional translator (a different person), to translate the English sentences into WAPE following the BBC style of writing.


\autoref{tab:dataset} provides the details of our new \warri dataset, containing \textbf{single-way parallel sentences} (translated  from WAPE BBC to English by two native speakers), and \textbf{multi-way parallel sentences} (where the English sentences have parallel translations in both WAPE BBC and Naija Wikipedia). Our test set composed of $500$ sentences and the remaining five sentences were for few-shot/in-context learning for LLMs. 

\paragraph{Other datasets in Naija} There are also other parallel translation datasets (i.e., Naija-English). We also perform an analysis and evaluation on them, and compare them to our \warri dataset.  \autoref{tab:examples} provides one example each per dataset. 

\begin{table}[t]
 \begin{center}
 \footnotesize
 \scalebox{0.90}{
  \begin{tabular}{lp{65mm}}
    \toprule
    \textbf{Lang.} & \textbf{Sample Sentences (English \& Pidgin) } \\
    \midrule
\textbf{\texttt{Bible}} \\
English &  And the Word became flesh, and dwelt among us \\
  Naija 1 & Den di Word kon shange to pesin and e stay with us for dis world \\
  Naija 2 &  Di W\d{o}d k\d{o}m bik\d{o}m human bin an Im liv wit \d{o}s \\
\midrule
\multicolumn{2}{l}{\textbf{\texttt{JW300}}} \\
 English &  What can we do to make wise use of our freedom?\\
 Naija  & Wetin go help us use our freedom well? \\
\midrule
\multicolumn{2}{l}{\textbf{\texttt{MAFAND}}} \\
 English &  Each group is supposed to submit its needs\\
 Naija  & Each group suppose bring di things wey dem need kom\\
 \midrule
 \multicolumn{2}{l}{\textbf{\texttt{UD}}} \\
 English &  And I love the job with all my heart\\
 Naija  & And I love di job as in wit all my heart \\
 \midrule
 \multicolumn{2}{l}{\textbf{\texttt{BBC}}} \\
 English & It is great - nothing is better than proving people wrong \\
 WAPE  & E dey great - nothing better pass make you prove pipo wrong \\

 \multicolumn{2}{l}{\textbf{\texttt{Wikipedia}}} \\
 English & He married one wife with 7 children. \\
 Naija & Na one wife im mari an dem don bon 7 pikin. \\
\bottomrule
  \end{tabular}
  }
  \vspace{-2mm}
  \caption{\footnotesize Example of different styles of Pidgin used in different corpora}
  \label{tab:examples}
  \end{center}
  \vspace{-2mm}
\end{table}

\begin{enumerate}[label=(\alph*)]
\item Bible: We found two Naija Bibles online. The first one was translated by Wycliffe Bible Translators, and is part of the freely available eBible corpus~\citep{Akerman2023TheEC}. Naija Wikipedia contributors also agree that this Bible conforms with Naija rather than WAPE. The other Bible translation was created by the Mercy Christian Ministry International (MCMI),~\footnote{\url{https://nigerianpidgin-bible.yolasite.com/}} which is written to be closely similar to an African languages in Nigeria including the use of underdot diacritics as they are used in  the Igbo language (e.g., ``W\d{o}d'' for ``Word''). \autoref{tab:examples} shows the two Naija Bible styles but there is not an established standard. We focus on our analysis on Wycliffe Bible (Naija 1) since the MCMI Bible (Naija 2) does not have the complete Bible online. We divided the Wycliffe Bible data into 31,051/1,500/1,500 TRAIN/DEV/TEST split. 

\item JW300: Similar to the Bible, JW300~\citep{agic-vulic-2019-jw300} is based on religious texts, bible studies and missionary reports of Jehovah Witness ministry in various languages. JW300 covers 343 languages including Naija. We divided the data into 23,322/ 1,500/ 1,500 TRAIN/DEV/TEST split. 

\item UD-Pidgin: This is based on the Universal Dependecy (UD) project for Naija~\cite{caron-etal-2019-surface}. The data is based on the transcript of a conversationbetween two Naija speakers.~\footnote{\url{https://github.com/UniversalDependencies/UD_Naija-NSC}} We divided the data into 6,241/ 1,500/ 1,500 TRAIN/DEV/TEST split. 

\item MAFAND: This is based on the news domain. The news articles were obtained from English Daily Trust newspaper (published in Nigeria), and translated to Naija~\cite{adelani-etal-2022-thousand}. We make use of the same split as the MAFAND corpus with 4,790/1,484/1,564. Unlike the other datasets, it can be considered as ``general domain'' similar to Wikipedia. 

\end{enumerate}

Aside from parallel corpora, large amounts of WAPE unlabelled texts have been collected in literature~ from BBC~\cite{ogueji-etal-2021-small} to train language models such as AfriBERTa. The AfriBERTa corpus has more than 160,000 sentences. Other sources of data for Naija are often smaller (e.g., Naija tweets~\cite{muhammad-etal-2022-naijasenti}). However, we primarily focus on the parallel data sources for our analyses.

\section{Experimental setup}

\begin{table*}[t]
 \begin{center}
 \footnotesize
 \scalebox{0.95}{
  \begin{tabular}{l|rrrr|rrr}
    \toprule
      &  & & & & \textbf{single-way} & \multicolumn{2}{c}{\textbf{multi-way}} \\
    \textbf{Metric} & \textbf{Bible}& \textbf{JW300}& \textbf{UD} & \textbf{MAFAND}& \textbf{BBC(WAPE)} & \textbf{Wiki (WAPE)} & \textbf{Wiki (Naija)} \\
    \midrule
\multicolumn{3}{l}{\textit{Jaccard Similarity ([0,1] range)}} \\
Unigram $\uparrow$ & 0.133 & 0.295 & 0.537 & 0.554 & 0.712  & 0.802 & 0.517 \\
Bigram $\uparrow$ & 0.025 & 0.086 & 0.149 & 0.178 & 0.289 & 0.371 & 0.167 \\
Trigram $\uparrow$ & 0.002 & 0.025 & 0.055 & 0.076 & 0.151 & 0.207 & 0.084  \\
\midrule
Levenshtein distance $\downarrow$ & 88.5 & 56.5 & 30.3 & 58.0 & 26.6 & 21.6 & 53.6  \\
\midrule
BLEU $\uparrow$ & 0.8 & 11.2 & 16.8 & 20.9 & 36.1 & 46.8 & 23.4   \\
ChrF++ $\uparrow$ & 20.5 & 30.4 & 43.7 & 54.7 &  65.2 &  73.4 & 51.3  \\
BERTScore $\uparrow$ & 72.4 & 79.6 &  82.6 & 82.4 & 87.4 & 90.5 & 79.8  \\
    \bottomrule
  \end{tabular}
}
  \vspace{-2mm}
  \caption{\textbf{Lexical overlap and Levenshtein distance on \warri benchmark}. Lexical overlap is measured by Jaccard similarity between English and WAPE (BBC) and Naija (Wikipedia). For multi-way \warri corpus, the source of the data is from Wiki. The WAPE translation is denoted as Wiki (WAPE) but the original text is in Naija. } 
  \label{tab:data_stats}
  \end{center}
  \vspace{-3mm}
\end{table*}

We conduct three types of experiments to find out if WAPE and Naija are similar to each other: (1) Statistical analyses of the texts obtained from different datasets to measure their similarity to English and to each other.  
(2) Cross-corpus zero-shot transfer results when an  MT model is trained on one dataset and evaluated on another. We expect domains that are similar should have a higher performance~\cite{adelani-etal-2021-effect,lee-etal-2022-pre}. Similarly, we expect transfer results to be higher if the pidgins are similar in terms of writing. 
(3) Prompting an LLM to find out whether WAPE or Naija is represented in Generative AI. We compare the results to the evaluation of \warri MT benchmark dataset when trained on \mafand.  ~

\subsection{Statistical analysis of the texts} First, we compute the lexical similarity between the English portion of each dataset and Pidgin by measuring \textbf{Jaccard similarity} (in percentage) for each corpus unigram, bigram, and trigram tokens. 
Secondly, we compute the \textbf{Levenshtein distance}~\citep{Levenshtein1965BinaryCC} which is an edit distance between each English test sentences and their translations to WAPE and Naija. Finally, we make use of three additional text generation metrics to measure their similarity to English: BLEU~\citep{papineni-etal-2002-bleu}, ChrF++~\citep{popovic-2017-chrf} and BERTScore~\citep{Zhang2020BERTScore}. BLEU and ChrF++ are n-gram matching metrics, while \textbf{BLEU} focuses on word-level matching, \textbf{ChrF++} helps with evaluating character-level differences which are more common for Pidgin. Therefore, it is more reliable. \textbf{BERTScore} is an embedding-based metric that measures the semantic relationship between the sentence embeddings of two sentences. Therefore, it has better correlations with the human judgments.  


\subsection{Cross-corpus zero-shot transfer results} 
We evaluate the performance of training an MT model on a source corpus and evaluate the performance on a target corpus. The source corpus are MAFAND, Bible, JW300, and UD, while the target corpus can be one of source corpora, and the \warri dataset i.e. single-way and multi-way test sets. We perform an evaluation based on ChrF++ due to its reliability about capturing the character-level differences between Pidgin and English. Following \newcite{adelani-etal-2022-thousand}, we leveraged a pre-trained model to train an MT model by fine-tuning M2M-100 (418M) on each source data, and evaluated on the remaining test sets of our datasets. 

\subsection{Prompting of LLMs} We prompted \gpt~\footnote{\gpt pre-training data is up to December 2023.} and \llama \& \textsc{70B}~\cite{Dubey2024TheL3}
to generate translations in either Pidgin or English in both zero-shot or few-shots settings (with one or five examples). A sample prompt is provided 
in \autoref{appendix_prompt}. The prompting result is compared with the supervised training of MT models on the \mafand dataset which is also in the general domain.

\section{Experimental Results}

\subsection{Statistical analysis results} 
In \autoref{tab:data_stats}, by computing a lexical similarity between the $n$-gram tokens of each genre, we show that WAPE in both news (\bbc) and Wikipedia domains consistently have a \textit{higher} Jaccard similarity score with its parallel English corpus for all $n$-grams, compared to other datasets with the Naija label. For example, the unigram similarity score for WAPE was around $0.712-0.802$ while the others are much lower between $0.133$ (Bible) and $0.554$ (\mafand). \ud, \mafand and \warri \wiki data sets have similar Jaccard similarities. 

Furthermore, Levenshtein distance provides an additional evidence of a difference between WAPE and Naija. It takes more than twice edit-distance to transform the English sentences to Naija (\wiki) than to WAPE (\bbc) and WAPE (\wiki). Naija (\wiki) requires more edits in characters, which shows that it is farther from English compared to the WAPE. In other words, these two pidgins are quite different than each other linguistically. Similarly, we find longer Levenshtein distance for other datasets: \jw \mafand, and \bible with $56.5$, $58.0$ and $88.5$ respectively. On the otherhand, \ud dataset has a shorter Levenshtein distance compared to others which we attribute to the shorter utterances of the dataset (see \autoref{tab:dataset}). 

Finally, our experiments on text generation metrics (e.g., BLEU, ChrF++ and BERTScore) show that WAPE (\bbc) is more similar to English than any of the other Pidgin datasets we evaluated. We find BLEU to be less reliable for this evaluation, achieving only $0.8$ for the \bible while ChrF achieve relatively higher scores. We attribute this result to several character-level differences between the Bible Pidgin and the English. In general, we find \textit{higher} scores for both WAPE (\bbc) and WAPE (\wiki) ($65.2 - 73.4$ ChrF++) than Naija (\wiki) ($51.3$). BERTScore evaluation also confirmed this finding by reaching to a score of $90.5$ for WAPE and $79.8$ for Naija.

\begin{figure}[t]
    \centering
    \includegraphics[width=1.02\linewidth]{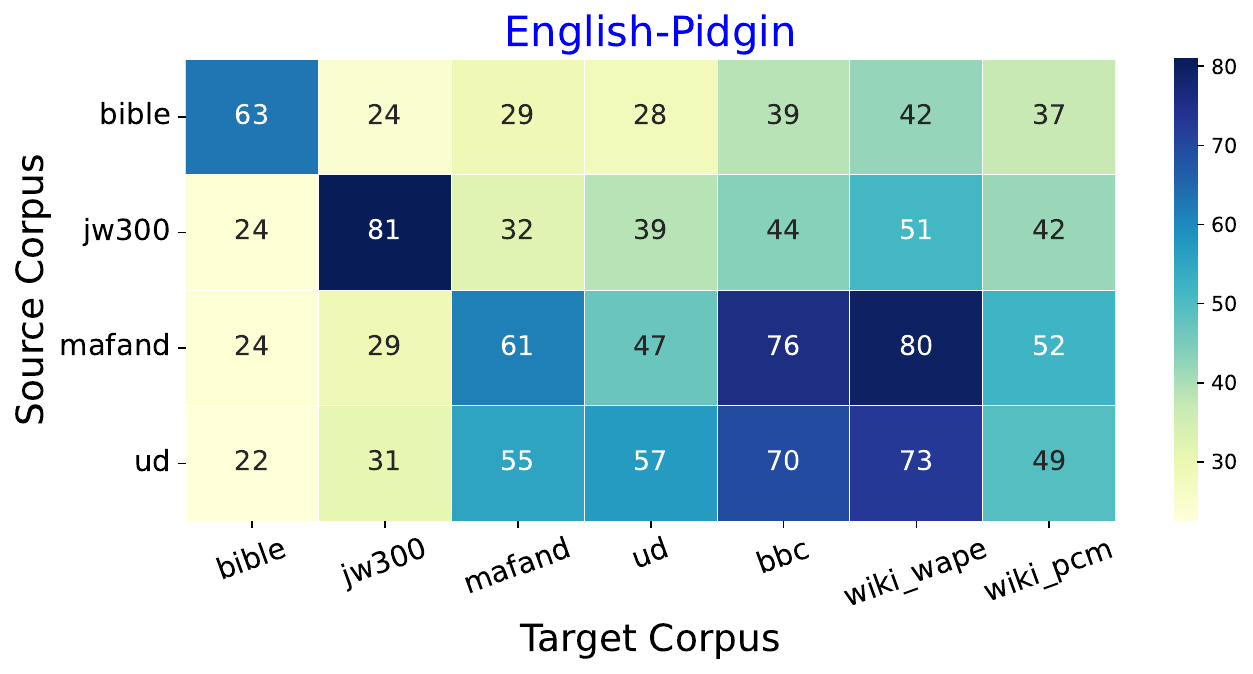}
	\includegraphics[width=1.02\linewidth]{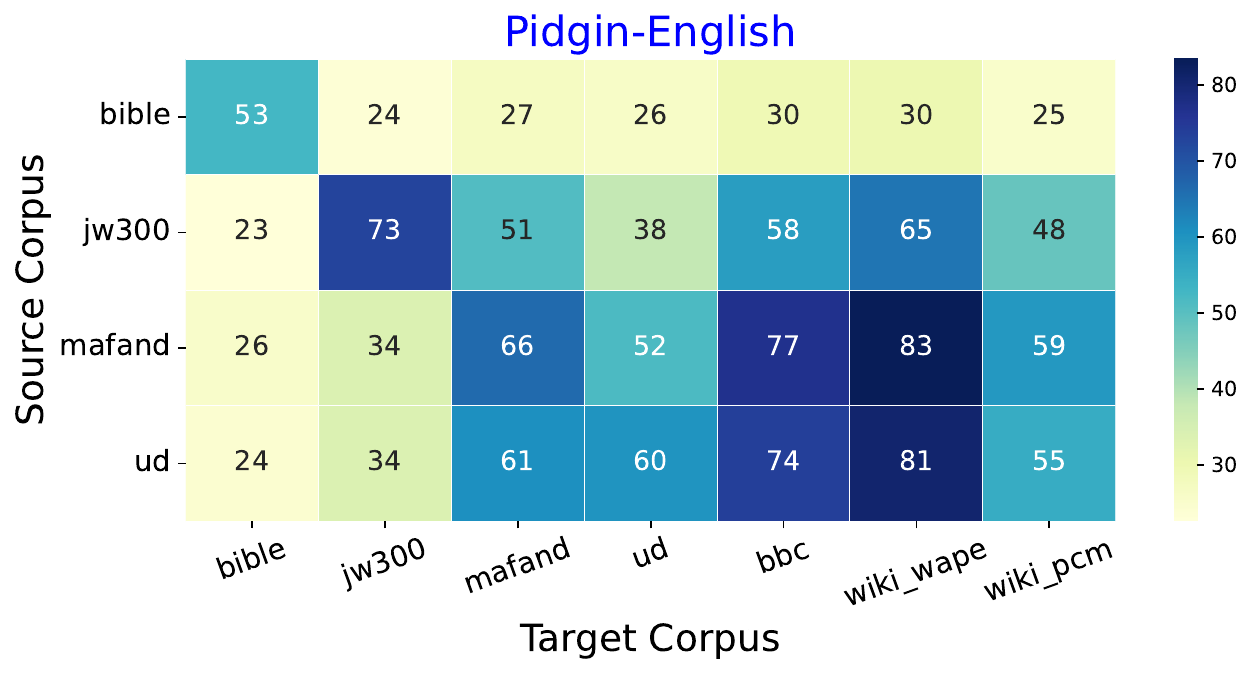}
    \vspace{-4mm}
    \caption{\textbf{Cross-corpus transfer results}: Evaluation based on ChrF++}
    \label{fig:transfer_en_pcm}
\end{figure}

\begin{table*}[t]
 \begin{center}
 \footnotesize
 \scalebox{0.86}{
  \begin{tabular}{l|rrr|rrr|rrr}
    \toprule
 & \multicolumn{3}{c|}{\textbf{Single-way (news)}} &  \multicolumn{6}{c}{\textbf{Multi-way parallel (Wiki)}} \\
    & \multicolumn{3}{c|}{\textbf{WAPE (BBC)}} & \multicolumn{3}{c|}{\textbf{WAPE (Wiki)}} & \multicolumn{3}{c}{\textbf{Naija (Wiki)}}\\
    \textbf{Evaluation Task} & \textbf{BLEU} & \textbf{ChrF++} & \textbf{BERTScore} & \textbf{BLEU} & \textbf{ChrF++} & \textbf{BERTScore} & \textbf{BLEU} & \textbf{ChrF++} & \textbf{BERTScore}\\
    \midrule
   \multicolumn{3}{l}{\textbf{\texttt{wape/pcm $\rightarrow$ en}}} \\
0-shot: \mafand $\rightarrow$ \warri & 57.6 & 76.3 & \textbf{94.5} & \textbf{68.6} & 83.4 & 96.2 & 35.0  & 59.1 & 86.7\\
0-shot: \llama & 51.5 & 74.8 & 92.1 & 58.5 & 79.5 & 93.6 & 37.8 & 64.5 & 89.0 \\
0-shot: \llamaB & 56.1 & 76.4 & 93.5 & 63.6 & 81.6 & 94.3 & 42.7 & 67.7 & 90.7\\
0-shot: \gpt & \textbf{59.3} & \textbf{78.8} & 94.4 & 65.5 & \textbf{83.6} & \textbf{96.3} & \textbf{43.5} & \textbf{68.9} & \textbf{92.2}\\


\midrule
 \multicolumn{3}{l}{\textbf{\texttt{en $\rightarrow$ wape/pcm}}} \\
0-shot: \mafand $\rightarrow$ \warri & 54.7 & 75.2 & 91.6 & 61.0 & 79.5  & 92.9 & 26.5 & 51.8 &  83.6\\
\addlinespace
0-shot: \llama  & 41.3 & 66.8 & 88.4 & 45.0 & 68.5 & 89.9 & 22.9 & 48.0& 81.9 \\
1-shot: \llama  & 41.0 & 67.9 & 87.9 & 44.0 & 72.1 & 87.7 & 26.1 & 50.2 &  82.3\\
5-shot: \llama  & 49.2 & 72.0 & 90.2 & 53.4 & 76.4 & 91.2 & 26.4 & 50.7 & 83.2 \\

\addlinespace
0-shot: \llamaB  & 46.1 & 69.8 & 89.7 & 43.8 & 67.8 & 89.5 & 25.4 & 50.9 &  83.2 \\
1-shot: \llamaB  & 50.6 & 73.1  & 90.8& 56.0 & 76.9 & 92.0  & 29.0 & 53.4 &  84.2\\
5-shot: \llamaB  & 58.1 & 77.2 & 92.1& 61.5 & 80.3 & 93.2& 28.0 & 53.1 & 84.7 \\

\addlinespace
0-shot: \gpt  & 51.8 & 72.3 & 91.4 & 53.8 & 74.8 & 92.0 & 26.7 & 51.7  & 83.1\\
1-shot: \gpt  & 58.7 & 76.9 & 92.6 & 57.7 & 79.7 & 92.8 & 29.6 & 54.3 & 84.6 \\
5-shot: \gpt  & \textbf{63.5} & \textbf{79.6} & \textbf{93.2 }& \textbf{64.9} & \textbf{83.1} & \textbf{93.8} & \textbf{30.0} & \textbf{54.7} &  \textbf{85.1} \\
    \bottomrule
  \end{tabular}
}
  \vspace{-2mm}
  \caption{\textbf{Evaluation on \warri dataset: single-way and multi-way parallel (same sentences translated to both pidgins) test sets}: We compared the performance of MT to different genres using GPT-4-Turbo and adapted M2M-100 (418M) from \mafand training set.}
  \label{tab:results}
  \end{center}
  \vspace{-3mm}
\end{table*}

\subsection{Machine translation evaluation}
While statistical analysis already proves the linguistic difference between WAPE and Naija, evaluation on a task provides additional perspectives (i.e., Would a model trained on WAPE, perform well on Naija, and vice versa? What is the transfer performance of a model trained on one pidgin to another for text generation tasks?).

\autoref{fig:transfer_en_pcm} shows our results on MT. The transfer performance depends on both the similarity of the domains (e.g. religious vs. news) and similarity of the pidgin style of writing. For example, religious datasets (e.g., \bible and \jw) generally transfer poorly to other domains. Similarly, \mafand and \ud in the news and spoken conversation domains also do not transfer well to religious domains. 

Both \mafand and \ud datasets often have a higher zero-shot transfer result to WAPE (\bbc and \wiki) than to Naija (\wiki). In the English-Pidgin direction, \mafand achieved between $76 - 80$ ChrF++ on WAPE while achieving only $52$ ChrF++ on Naija (PCM). Surprisingly, we find \mafand transfering better to WAPE than to its own test set showing the simplicity of generating WAPE compared to Naija. We have a similar observation when transfering from UD. Our evaluation on the Pidgin-English also confirms this hypothesis that translating from WAPE to English is easier for MT systems than Naija. 

\subsection{LLM performance on \warri Benchmark}
In this section, we focus on finding out which pidgin is represented in the current LLMs, and whether they support several pidgin variants which are accessible for different communities of Naija speakers. We evaluated the performance of LLM in translating from and into WAPE and Naija. 

\paragraph{\mafand MT model and LLMs represent WAPE more}
\autoref{tab:results} shows the result of evaluation of the \warri MT results. In the direction of \textbf{wape/pcm$\rightarrow$en}, adapting \mafand MT model to WAPE gave an impressive result in both single-way ($76.3$ ChrF++) and multi-way parallel ($83.4$ ChrF++) scenarios. However, the performance on Naija (\wiki) is much worse ($-24.3$ drop in ChrF++). This shows that the fine-tuning corpus most likely represents the WAPE. Similar observation was found in \gpt and \llama \& \textsc{70B} evaluation, although the performance of the latter was worse especially on Naija. Similarly, for the \textbf{en$\rightarrow$wape/pcm}, in zero-shot setting, \mafand MT model gave the best performance over \gpt and \llama on WAPE in zero-shot setting, and competitive performance on the Wikipedia genre ($51.8$ ChrF++) compared to \gpt ($51.7$ ChrF++).

\begin{table*}[t]
 \begin{center}
 \footnotesize
 \resizebox{\textwidth}{!}{%
  \begin{tabular}{l|l}
    \toprule
    \textbf{Method} & \textbf{Translation} \\
    \midrule
English &  Rimi said he gave them the money because they did all that was expected  of them to do in their different areas.  \\
Reference (Naija) &  Rimi sey im giv dem di moni bikos dem "folo do evritin wey dem sopos du for dia difren aria. \\
Reference (WAPE) & Rimi tok say im give dem di money becos dem do evritin wey dem suppose do for dia different areas. \\
\midrule
\multicolumn{2}{l}{\texttt{English-Naija translation}} \\
0-shot: \gpt &  \textcolor{cyan}{Rimi talk say e give dem the money} \textcolor{red}{because} dem do \textcolor{red}{everything} wey dem \textcolor{cyan}{suppose do for} \textcolor{red}{their different areas}.

\\\

1-shot: \gpt &  \textcolor{cyan}{Rimi tok say im give} \textcolor{violet}{dem di moni} \textcolor{red}{because} dem do all di tin wey dem \textcolor{cyan}{suppose do for dia different area}.  \\

5-shot: \gpt &  \textcolor{violet}{Rimi tok se im dash dem di moni} \textcolor{cyan}{becos} dem do all di tins wey dem \textcolor{red}{suppose} do for \textcolor{violet}{dia diffren} \textcolor{red}{areas}.  \\

\addlinespace
0-shot: \llamaB &  \textcolor{cyan}{Rimi talk say him give dem money} \textcolor{red}{because} dem do all wetin dem \textcolor{cyan}{suppose do for dia different area}. \\
1-shot: \llamaB &  \textcolor{cyan}{Rimi tok say im give dem} \textcolor{violet}{moni} \textcolor{red}{because} dem do all wet dem \textcolor{cyan}{suppose do for dia different areas.}  \\
5-shot: \llamaB &  Rimi \textcolor{red}{talk} say \textcolor{violet}{im dash dem di moni bikos} dem do wetin dem \textcolor{cyan}{suppose do for dia different areas}.  \\

\midrule
\multicolumn{2}{l}{\texttt{English-WAPE translation}} \\
0-shot: \gpt &  Rimi  \textcolor{red}{talk} say e give dem di money  because dem do all wey dem suppose do for their different areas. \\
1-shot: \gpt &  Rimi  \textcolor{red}{talk} say e give dem the money  because dem do all wey dem expect make dem do for their different areas. \\
5-shot: \gpt &  Rimi  \textcolor{red}{talk} say e give dem di money  because dem do all wey dem expect dem to do for their different areas. \\

\addlinespace
0-shot: \llamaB &  Rimi tok say \textcolor{red}{him} give dem money because dem do all wetin dem suppose do for dia different areas.  \\
1-shot: \llamaB &  Rimi \textcolor{red}{talk} say im give dem moni because dem do all wetin dem suppose do for dia different areas.  \\
5-shot: \llamaB &  Rimi \textcolor{red}{talk} say im give dem di money because dem do all wet dem expect dem to do for dem different areas.  \\

    \bottomrule
  \end{tabular}
  }
  \vspace{-2mm}
  \caption{\footnotesize \textbf{Qualitative analysis on Predicted translations on \warri dataset:} multi-way parallel output. Words/phrases expressed in Naija are in \textcolor{violet}{violet color}, WAPE words are in \textcolor{cyan}{cyan}, while English words that ought to be translated are in \textcolor{red}{red}.}
  \label{tab:qualitative_results}
  \end{center}
  \vspace{-2mm}
\end{table*}

\paragraph{Can we teach LLMs different genres with only a few examples?}
Our result (Table 4) of prompting \gpt, \llama and \llamaB shows that providing one or five examples is effective for extra performance boost to generate generating Pidgin sentences.\footnote{The exact 1-shot and 5-shots examples are provided in the \autoref{sec:five_shots}} For \gpt, the performance improved over zero-shot result by $+4.9$ ChrF++ when the LLM is prompted with one example translation of WAPE, and $+8.3$ ChrF++ when prompted with five examples, on the multi-way test set. However, the boost in performance is very small when Naija (\wiki) examples are provided. It is only $+2.6$ and $+3.0$ when one example and five examples are provided during the prompting of \gpt. This shows that \gpt is more biased toward the WAPE than Naija and it is difficult to teach the LLM with few examples. The reason for this performance difference is because the WAPE (BBC) is the largest unlabelled data available on the web~\citep{ogueji-etal-2021-small}. Other sources that are more representative of Naija, are often in smaller quantity (e.g., Wikipedia). We observe a similar trend for the LlaMa models where \llamaB attained up to $80.3$ ChrF++ with 5-shots ($-2.7$ points when compared to \gpt), while \llama achieved $76.4$. 

We provide a qualitative example in \autoref{tab:qualitative_results}, where we show that with one or five examples, the \llamaB LLM slightly changes its writing style to be more similar to Naija but sometimes the model combine the vocabulary of WAPE and Naija which leads to misunderstandings. For example, in the 5-shot translation of ``...was expected of them to do in their different areas'', \llamaB translated it to be \textbf{``suppose do for dia different areas''} which is more similar to the WAPE translation of the same sentence. However, in Naija the words like ``suppose'', ``different'' and ``area'' are spelled differently (e.g., \textbf{``sopos du for dia difren aria''}). On the other hand, \gpt produced an (almost) accurate translation into Naija except the use of ``suppose'' rather than ``sopos'' and ``becos'' (a WAPE word) instead of ``bikos''. This implies that the model is able to learn in-context. However, it is still biased towards WAPE without few-shot examples. With more examples, we may be able to teach the model Naija with supervised fine-tuning of the instruction data containing Naija-English parallel sentences. Qualitative examples for WAPE show that the LLMs are able to generate sentences correctly in zero-shot setting without additionally few shot examples which confirms our hypothesis that the LLMs are biased towards WAPE.

\section{Qualitative interviews with Naija Wikipedia contributors}
\label{sec:interviews}

To validate our study, we interviewed two native speakers of Naija who contribute to the writing and editing of Naija Wikipedia articles. Some Wikipedian contributors have online public profiles with links to their email addresses and social media accounts (e.g, Twitter or Linkedln). We sent emails to two Naija contributors with the online public profiles and conducted interviews ($\sim1$ hour each) with each of them.~\footnote{We provided honorarium of $\$11$ to each interviewee.} 

Our first observation is that the Naija Wikipedia contributors are not linguists or language experts but they are volunteers without a formal linguistic training. They have a passion for Naija and make an effort toward establishing a writing system which is very similar to the way it is spoken in their community (within Nigeria), rather than targeting a wider West African audience like BBC. They make efforts to create a standardized way of writing during the Naija Wikipedia incubator program. Through these efforts, Naija is included as a separate language on Wikipedia. The volunteers were part of the Wikipedia incubator program from the start, and they are part of the editors team of Naija Wikipedia. This team mentors new contributors about how to write the Naija reflecting the patterns how Naija is spoken (sometimes with a few adjustments to make it readable since original spoken Naija form could be different than English (e.g. ``moto'' instead of ``car'')).  

The volunteers also mentioned that they consult the available literature (e.g., \citet{Ofulue_guide}) about Naija and follow the recommended rules by Naija linguists ~\cite{Balogun2013InDO,Aghoghovwia_meeting} before starting to contribute to Naija Wikipedia. 

In terms of content of the Naija entries on Wikipedia, the contributors focus on the biographies of notable people (e.g., musicians and actors) in Nigeria. They are not allowed to contribute to sensitive topics (e.g., health) except when it is a direct translation from an high-resource language (e.g., English). To achieve this, they prefer words that come from local Nigerian languages (e.g., Hausa, Igbo and Yoruba), which many Nigerians are familiar with rather than words that are commonly understandable across West Africa. In general, the interviews with the Naija Wikipedia contributors confirm our results that they follow some convention distinguishing them from the WAPE writing convention.

\section{Conclusion}
Different versions of Pidgins mixed with English and local languages are used in West Africa but not all of them have standardized writing systems. Since 2017, BBC broadcasts (on Internet) in WAPE target the West African countries with the goal of reaching a wider audience across countries in a standardized writing style. 
 
Nigeria is a multilingual country with both richness and challenges that come along with the linguistic diversity. Although the official language is English, Naija is a lingua franca that brings speakers of different Nigerian languages together regardless of their linguistic, social or educational backgrounds. Since 2022, it is also a written language on Wikipedia.





Although both pidgin varieties are used in Nigeria, we prove that WAPE and Naija are different from each other linguistically and current Generative AI models are built upon WAPE only. This is probably due to more availability of data on the Internet for the WAPE rather than Naija. 

Lack of data on low resource languages is a key challenge for current AI systems. In our paper, we show that the situation is much more challenging for linguistically rich areas (e.g., West Africa). More specifically, pidgin varieties with the most data on the Internet gets represented on AI systems and the others may not be visible. This could potentially lead to a bias towards favoring language preferences of certain speakers/users instead of being more inclusive toward the users/speakers of other pidgins. Although our analysis focuses on Naija spoken in Nigeria, we hope to extend our analysis to other English-based pidgins in West Africa (e.g., Ghananian Pidgin, Cameronian Pidgin, and Krio in the future) as well. 



\section{Limitation}
There are few limitations of our work (1) Our evaluation dataset is small, although we argue that 500 may be good enough as a test set for MT. However, we only have a maximum of 5 sentences we could use for the few-shot learning or in-context learning. Moreover, with additional sentences (e.g. 2.5K-5K parallel sentences as recommended in \citep{adelani-etal-2022-thousand}), we may be able to adapt M2M-100 model to produce better generation of the Wikipedia genre. (2) Our analysis is limited to one task which is machine translation, we hope to extend this analysis to other tasks in the future as well. 


\section*{Acknowledgment}
We are grateful to Sergiu Nisioi and John Pavlopoulos for their time and feedback on an early draft of the paper. We thank OpenAI for granting the API credits (through their Researcher Access API program to Masakhane) enabling the evaluation of GPT-4 LLMs. We thank the volunteer Wikipedia contributors for Naija for their insights and explanations, anonymous reviewers, area chairs and senior area chairs of NAACL25 for their feedback and recommendations on our paper. 

\bibliography{custom}

\appendix

\section{Prompt Template}
\label{appendix_prompt}

\section{MAFAND training}
We fine-tune MAFAND dataset on M2M-100 (418M) using the same hyparameters stated in \citet{adelani-etal-2022-thousand} i.e. number of training epochs of 10, batch size of 32, source and target maximum sequence length of 200, and beam size of 10. 

\section{Licence of \warri}
We plan to release it publicly under the CC-4.0-NC due to BBC portion of the dataset that cannot be for commercial use. However, \warri (multi-way) has a licence of CC-4.0 international. 


\section{Five shot examples provided}
\label{sec:five_shots}

\begin{table*}[t]
\begin{center}
\resizebox{\textwidth}{!}{%
\begin{tabular}{lp{145mm}}
\toprule
\textbf{}
& \textbf{Prompt} \\
\midrule
\textbf{Task Description} & You are a helpful assistant who is an expert in translating English sentences to Pidgin using two varieties: West African Pidgin English and Nigerian-Pidgin, I would provide you with five examples of the  different varieties, your task is to follow the style of the writing of the specified variety when translating the sentences. \\
\textbf{Example}  & \textbf{Example 1:} \\
& \textbf{English}: Innocent Ujah Idibia was born on 18 September 1975, that is well known as 2baba, a Nigerian singer, songwriter, producer, philantropist. \\
& \textbf{West African Pidgin English}: Innocent Ujah Idibia wey dem born for 18 September 1975, wey dem know as 2baba, be a Nigerian singer, songwriter, producer, philantropist. \\
& \textbf{Nigerian-Pidgin}: Innocent Ujah Idibia (dem bon am for 18 September 1975), wey pipul no wel wel as  2baba, na Naija singa, songraita an podusa an im sabi dash pipul moni an gift wel wel. \\
\\

\textbf{Example}  & \textbf{Example 2:} \\
& \textbf{English}: He was born in Jos, Nigeria \\
& \textbf{West African Pidgin English}: Dem born am for Jos, Nigeria \\
& \textbf{Nigerian-Pidgin}: Dem bon am for Jos for inside Naija. \\

\\

\textbf{Example}  & \textbf{Example 3:} \\
& \textbf{English}: He is from the Idoma ethnic group \\
& \textbf{West African Pidgin English}: Im be from di Idoma ethnic group \\
& \textbf{Naija}: Im na Idoma pesin.\\
\\

\textbf{Example}  & \textbf{Example 4:} \\
& \textbf{English}: Idoma is in the southern part of Nigeria \\
& \textbf{West African Pidgin English}: Na southern part of Nigeria Idoma dey \\
& \textbf{Nigerian-Pidgin}: Idoma dey for di south side for Naija. \\
\\

\textbf{Example}  & \textbf{Example 5:} \\
& \textbf{English}: Before July 2014, he used 2face Idibia as his stage name \\
& \textbf{West African Pidgin English}: Before July 2014, i dey use 2face Idibia as im stage name \\
& \textbf{Nigerian-Pidgin}: Bifor July 2014 na 2face Idibia bi di nem wey im dey yuz for stej.\\
\\

\textbf{Prompt} & 'Translate this sentence to Nigerian Pidgin \\
\\
\textbf{Input}  & Alexander Abolore Adegbola Akande was born on 17 January 1980, well known as 9ice, a Nigerian singer, dancer, and songwriter. \\

\midrule
\textbf{Output:} & Alexander Abolore Adegbola Akande (dem bon am for 17 January 1980), wey pipul sabi well well as 9ice, na Naija singa, dansa, an songraita. \\
\bottomrule
\end{tabular}
}
\caption{\textbf{Prompt template used for MT}. An example prediction by \gpt}
\label{tab:prompt_templates}
\end{center}
\end{table*}

\end{document}